# Analytical Calculation of Weights Convolutional Neural Network

P.Sh. Geidarov

*Institute of Control Systems of the Ministry of Science and Education of Azerbaijan,*

**Abstract:** In this paper proposes an algorithm for the analytical calculation of convolutional neural networks without using neural network training algorithms. A description of the algorithm is given, on the basis of which the weights and threshold values of a convolutional neural network are analytically calculated. In this case, to calculate the parameters of the convolutional neural network, only 10 selected samples were used from the MNIST digit database, each of which is an image of one of the recognizable classes of digits from 0 to 9, and was randomly selected from the MNIST digit database. As a result of the operation of this algorithm, the number of channels of the convolutional neural network layers is also determined analytically. Based on the proposed algorithm, a software module was implemented in the Builder environment C ++, on the basis of which a number of experiments were carried out with recognition of the MNIST database. The results of the experiments described in the work showed that the computation time of convolutional neural networks is very short and amounts to fractions of a second or a minute. Analytically computed convolutional neural networks were tested on the MNIST digit database, consisting of 1000 images of handwritten digits. The experimental results showed that already using only 10 selected images from the MNIST database, analytically calculated convolutional neural networks are able to recognize more than half of the images of the MNIST database, without application of neural network training algorithms. In general, the study showed that artificial neural networks, and in particular convolutional neural networks, are capable of not only being trained by learning algorithms, but also recognizing images almost instantly, without the use of learning algorithms using preliminary analytical calculation of the values of the neural network's weights.

**1. Introduction and purpose of the work.** It is known that convolutional deep neural networks are currently widely used throughout the world [1, 2]. The popularity of these networks is due to the fact that deep convolutional neural networks show the best results, including in pattern recognition tasks. However, it is also known that the process of creating convolutional neural networks has a number of difficulties. Among which are:

1. Excessive duration of the neural network training procedure. For some tasks, the training time can last days, weeks, or months.
2. The need to have a huge database for training a neural network, amounting to tens and hundreds of thousands of data, and sometimes more. Such database is not always accessible to neural network developers.
3. Lack of accurate indications for choosing parameters of a convolutional neural network such as: kernel size, number of channels in convolutional layers, number of convolutional layers, presence of pooling channels, number of neurons in a fully connected neural network, etc.

To this end, this work proposes an algorithm for creating a computable convolutional neural network that will be able to identify images immediately - without the use of learning algorithms.

Let us recall that to improve the capabilities of artificial NNs of direct propagation, NN architectures of a multilayer perceptron based on metric recognition methods (NMRM) were proposed [3, 4], which implement algorithms of metric recognition methods, such as the nearest neighbor method, the nearest n neighbors method , potential method, etc., and at the same time have a multilayer perceptron architecture, which makes it possible to additionally train these networks like a regular multilayer perceptron. In [3, 4] it was shown that the weight values for these NNs can be calculated analytically using formulas using a small set of images, which made it possible to immediately obtain a workable NN without the use of learning algorithms and training data sets. Experiments have shown that the process of preliminary analytical calculation of the values of NN weights using formulas without training the NN is performed very quickly - in a fraction of seconds. At the same time, these neural networks are also capable of learning, where the process of training the resulting neural network turns into a process of additional training. Also, based on experiments [3], it was shown that the process of additional training of the NN occurs much faster than full training of the NN in the classical way - using the initial random generation of weight values. In addition, in experiments in [4] on a fully connected neural network, it was also shown that for computed fully connected neural networks, the additional training process requires a significantly smaller volume of the training data set.

It should also be noted that currently there are developed methods with preliminary calculations of weights, for example those given in [5-7]. But these methods do not calculate the exact values of the weights, but are variations of the smarter random initialization of the weights. Methods in [5-7] calculate not the values of the NN weights themselves, but the values of the range limits for random initialization of the weights in each convolutional layer. The purpose of using such initialization methods is aimed at solving the problem of training very deep networks with a very large number of layers (more than ten). It is known that in the process of training such deep neural networks, the outputs of neurons in the layers often form too large or, on the contrary, too small values. It is greatly slows down the training of the deep neural network and often leads the learning process to paralysis or to a local minimum [8-10]. The methods for initial initialization of weights in [ 5–7] make it possible to prevent this process and thus speed up the process of the deep neural network training algorithm. But these methods of initial initialization of weights are not capable of calculating a workable neural network without using a training algorithm. Algorithms and formulas for calculating the parameters of a multilayer perceptron proposed in [3, 4] make it possible to immediately obtain, without training, the values of the weights of a neural network, which becomes capable of immediately recognizing images without performing learning process.

Another type of neural network architecture for which weight values can be calculated is probabilistic neural networks, such as PNN networks (probabilistic neural network) [11] and GRNN networks (general regression neural network) [12-13]. For these neural networks, the weight values are calculated in one complete pass of the training set. This process is called training and requires a large amount of data, the amount of which is commensurate with the amount of the MNIST training dataset. In addition, PNN and GRNN networks cannot be trained using backpropagation algorithms, which does not allow these networks to achieve high results. Computational neural networks based on metric recognition methods, on the contrary, are classical perceptrons that can be trained using a backpropagation algorithm.

Convolutional deep neural networks have more complex architectures than a conventional fully connected perceptron. The ability to calculate a convolutional neural network will make it possible to show that convolutional deep neural networks are capable of not only learning with training data, but also immediately remembering recognized images without training. And since the architecture of artificial convolutional neural networks was originally borrowed from biological neural networks, this possibility can probably also explain why in biological living systems, training: animals, birds, people (children) does not require large databases [14].

It should also be noted that there are also learning algorithms designed for use with convolutional neural networks for cases where the training data sets are too small. In 2019, the One-shot learning algorithm was proposed, which can also be used for convolutional neural networks [15-19]. In this regard, it is necessary to note the significant differences between the "One-shot" algorithm and the one proposed in this work:

1) First of all, it should be noted that a neural network trained using the "One-shot" scheme is not a traditional classification algorithm. Such a neural network is not capable of recognizing more than two images (classes). Neural network implemented based on " One-shot" learning " has only one single output, which can take only two states - values >0.5 or <0.5 (1 or 0). Therefore, the " One-shot" algorithm learning " is used where it is necessary to recognize only two states (two classes), for example, to distinguish a man from a woman in a photograph, or to distinguish a cat from a dog, or to answer the question: does a person look like the photo in his passport or not. In all these cases, only two states are used, determined by the neural network - similar or not similar. But in tasks where it is necessary to recognize the number of classes more than two - the use of "One-shot" technology learning is no longer possible, including for the task of recognizing the MNIST database. The proposed technology for analytical calculation of weights creates a traditional one convolutional neural network that can recognize more than two images.

2) In addition, we also note that the "One-shot" technology does not analytically calculate the values of the weights of the neural network, as proposed in this article, but determines them by fully training the neural network using the backpropagation algorithm. In the learning algorithm "One-shot" three images are submitted to the input of the neural network: two images that are very similar to each other (anchor, positive) and one image (negative) that is not similar to the image being trained. If, for example, there are 20 images (10 from each class), then to train the neural network using the "One-shot" scheme learning it is required to create numerous combinations of three images (triplets) from 20 images. For the case with 20 images, we obtain a set of training data from a set of variants of three images equal to n = 10*9*10*2 = 1800. Using this database, the neural network is trained with multiple training epochs. Obviously, in terms of time, the creation of a neural network using the "One-shot" technology will take significantly more time than the analytical calculation of weights proposed in this work, which is calculated in fractions of a second or a minute.

3) Convolutional neural network architectures trained using the "One-shot" algorithm learning require too much memory. This is due to the fact that three images are supplied to the input of the neural network at once (anchor, positive, negative) and, accordingly, the neural network has three convolutional neural network, called siamese twins. This architecture makes the neural network cumbersome and requires significant memory resources. The technology proposed in this work for calculating the weights of a convolutional neural network does not have such a disadvantage, since a traditional single convolutional neural network is created.

Based on all of the above, we can say that the scientific novelty of this work lies in the fact that for the first time a convolutional neural network is created, for which all weight values are analytically calculated.

## 1. COMPUTING THE FIRST CONVOLUTIONAL LAYER

In Fig. 1 shows an example of a 28 by 28 pixel "0" image taken from the MNIST test database. In this algorithm, for all images, the values of the images of the array M [i][j] are brought into accordance with the condition if M [i][j]>127, then M [i][j] = 255, otherwise M [i] [j] = 0. In reality, this condition was not necessary, but in this implementation of the algorithm, to simplify the work, this condition was fulfilled (Fig. 1).

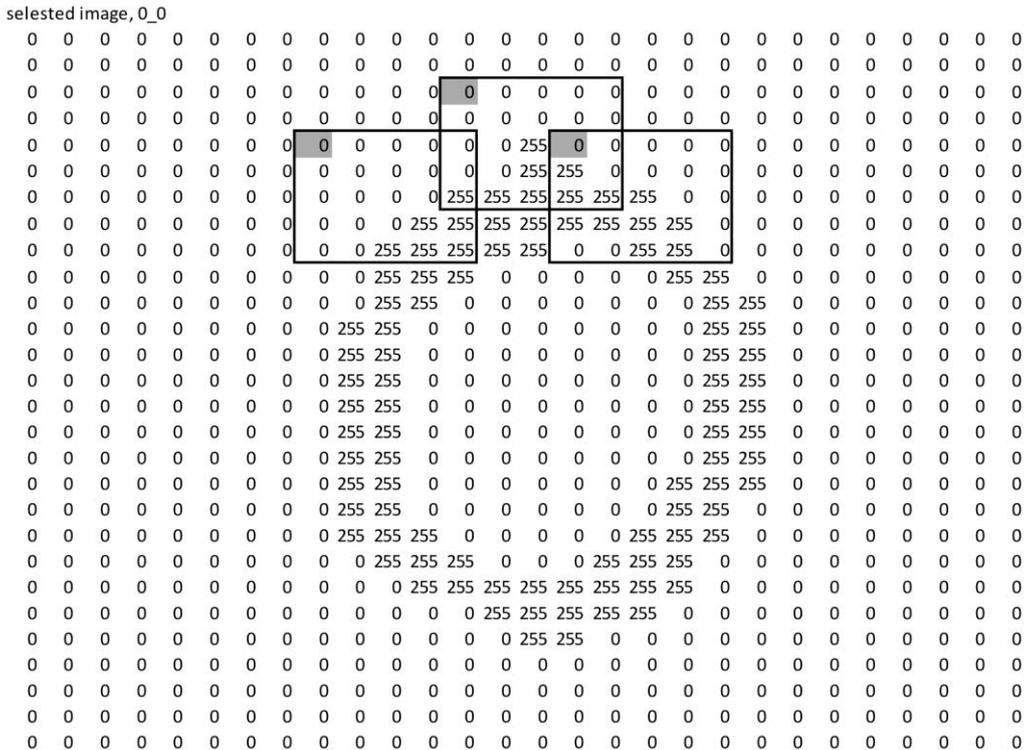

Fig.1. An example of some selected features on the image "0".

A 5 by 5 pixel array was chosen as the kernel dimension (Fig. 1). The kernels of the first convolutional layer were determined on the basis of a part of the array images with a dimension of 5 by 5 pixels, selected at the boundaries of the transition of the image from black to white pixels (Fig. 1). In this case, each time an array M1[5][5] was selected, in the center of which there is an array M2[2][2] in which there are both values: "0" and "255" (Fig. 2).

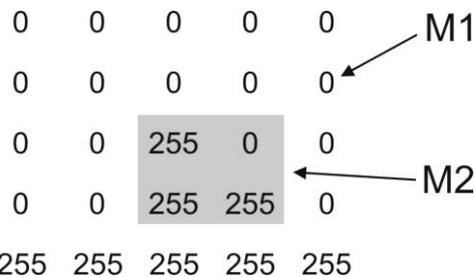

Fig. 2. A part of the image measuring 5 by 5 pixels, on the basis of which the channel kernel for the first convolutional layer is calculated.

Examples of such image arrays are shown in Fig. 1. Thus, the algorithm sequentially reviews all possible options on the selected 10 images with a step of 2 pixels and determines all matrices that meet this condition. The location of each found array is stored in a separate array, referred to below as the "common feature channel", in the form of numbers 1 (Fig. 3). The coordinates of which correspond to the coordinates of the upper left pixel of each found array (Fig. 1). Since this example only uses 10 selected images (1 image for each class), we end up with 10 feature channels. Figure 3 shows an example of a common feature channel for the selected image "0" in fig. 1. Accordingly, the dimension of the array of the general channel of features is less than the dimension of the image array by 4 units (28-(5-1) = 24) (Fig. 3). This table of the feature channel indicates the points (units) at which the image features are located, on the basis of which the channel kernels of the first convolutional layer are calculated. The coordinates of each point in the feature channel array corresponds to the initial coordinate of the image matrix (Fig. 1), on the basis of which the kernel of the first convolutional layer will be calculated.

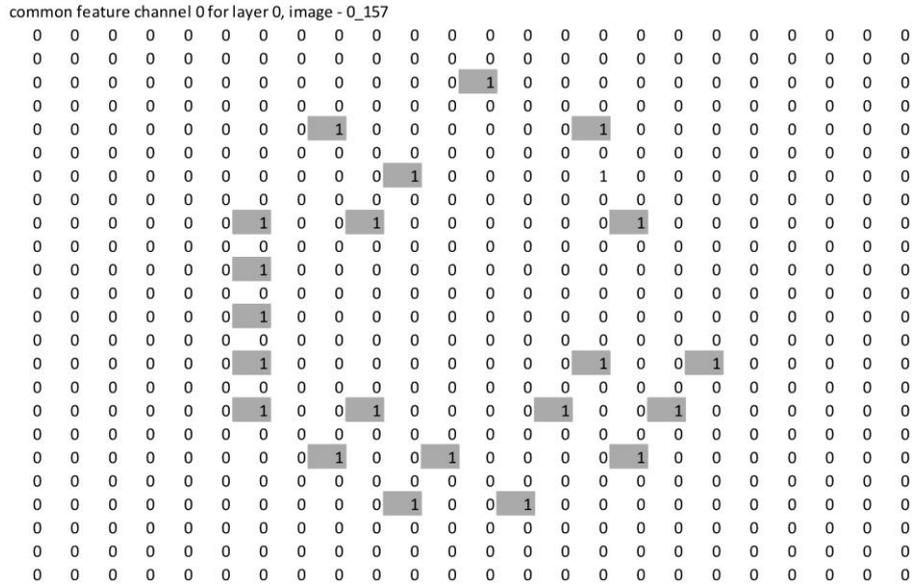

Fig. 3. Common feature channel for image "0", randomly selected in the MNIST test database with sequence number 157.

The value of each kernel is calculated based on a 5 by 5 image matrix, shown, for example, in Fig. 1, the initial coordinates of which are determined by the coordinates of the cells of the feature channel, the values of which are equal to 1. The kernel values for the first convolution layer are determined by the expression:

$$w_{ij} = \frac{p_{ij} \cdot 2}{\max(p_{ij})} - 1 \qquad (1)$$

Here $p_{ij}$ is the value of a 5 by 5 matrix (Fig. 4a), $w_{ij}$ – the value of the kernel weight of the first convolutional layer, $max(p_{ij})$ – the maximum pixel value of the selected image (in Fig. 1 $max(p_{ij})$ = 255).

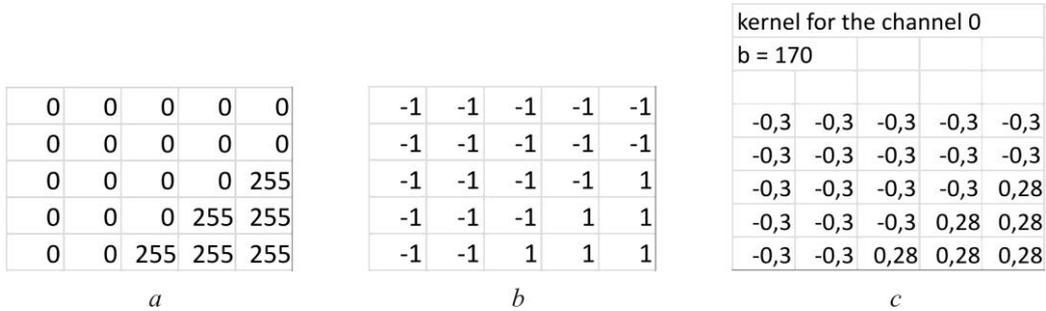

Fig. 4. Calculation of the kernel for the channel of the first convolutional layer, (a) - image feature, (b) - channel kernel without correction, (c) - kernel with corrections.

Expression (1) for a matrix with a dimension of 5 by 5 (Fig. 4a) determines the weight value $w_{ij} = -1$ for the value 0 and the weight value $w_{ij} = 1$ for the maximum value 255 (Fig. 4b). In general, if the image were in grey shades in the range [0, 255], then for all other values of the image pixels, according to expression (1), intermediate values $w_{ij}$ would be calculated, varying in the range from [-1, 1].

In this algorithm, the bias value $b$ for a given kernel are calculated from the expression:

$$b_{\ker nel}^{(layer\_1)} = \left( K \cdot \sum_{i=1}^{5} \sum_{j=1}^{5} (p_{ij} \cdot w_{ij}) \right) \Big/ 100 \qquad (2)$$

Here $K$ is the selected percentage value, $p_{ij}$ is the value of the image pixel in the image feature matrix (Fig. 4a).

Having the values of the kernel weights $w_{ij}$ and the bias $b$, the state values of the neurons of the first channel of the first convolutional layer are calculated using expression (3). The kernel matrix $\bar{M}_{kernel}$ is sequentially multiplied by all possible matrices of pixel values $\bar{M}_{kernel}$ with the same dimension (5:5) of the image in Fig. 1. In this case, for each new value $p^{(layer\_1)}_{channel\ value}$ for the image matrix, $\bar{M}_{kernel}$ the horizontal and vertical coordinates change sequentially with a change in the indices $k$ and $l$:

$$p^{(layer\_1)}_{channel\ value} = \bar{M}_{kernel} \cdot \bar{M}_{image} - b_{kernel} = \sum_{i=1}^{5}\sum_{j=1}^{5}\left(p_{k+i,l+j} \cdot w_{ij}\right) - b_{kernel} \quad (3)$$

Here in formula (3) $k$ is the vertical shift of the kernel in the image, and $l$ is the horizontal shift of the kernel in the image. These values will also correspond to the coordinates of the value $p^{(layer\_1)}_{channel\ value}$ in the channel of the first convolutional layer.

It should be noted that in order to exclude an uncontrolled increase in channel values $p^{(layer\_1)}_{channel\ value}$ when moving from one convolutional layer to another layer, it is necessary to first adjust the values of the kernel weights, so that all channel values $p^{(layer\_1)}_{channel\ value}$ would not exceed the maximum value in the previous convolutional layer (for the first layer, this maximum image pixel value is 255). To do this, with the obtained values of the kernel weights in Fig. 4b expression (3) calculates the value $p_{main\ channel\ value}$, on the basis of which the data of these weights were calculated. If, as a result, the value obtained from expression (3) $p^{(layer\_1)}_{main\ channel\ value}$ is greater or less than the value 255, then in this case the coefficient of increase or decrease $t$ is determined by the expression:

$$t = p^{(layer\_1)}_{main\ channel\ value} / 255, \quad (4)$$

Next, all values of the kernel weights are adjusted (Fig. 4c) taking into account this coefficient $t$ according to the formula:

$$w_{ij} = w_{ij}/t \quad (5)$$

$b$ is recalculated using expression (2). Thus, the values of the kernel weights and bias are determined for the first (zero) channel of the first convolutional layer. Moreover, if the value is $p^{(layer\_1)}_{main\ channel\ value} < 0$, then this feature is skipped in the feature channel and a transition is made to the next single value in the feature channel in Fig. 3.

Based on the results of the obtained values of weights and bias $b$ using expression (3), all values of the state of the neurons of the first (zero) channel of the first convolutional layer are finally calculated for the first recognized image "0" (Fig. 5), (hereinafter the "real" channel will be called). Moreover, in (3), the values of k and l change sequentially in a cycle by 1, which ensures the movement of the kernel across the image, according to the architecture of convolutional neural networks. And the neuron activation function (ReLu) is calculated by the expression:

$$if\ p^{(layer\_1)}_{channel\ value} > 0, mo\ f\left(p^{(layer\_1)}_{channel\ value}\right) = p^{(layer\_1)}_{channel\ value}$$
$$if\ p^{(layer\_1)}_{channel\ value} \leq 0, mo\ f\left(p^{(layer\_1)}_{channel\ value}\right) = 0 \quad , \quad (6)$$

In Fig. 5 you can see that the maximum value of the first real channel for the first selected image "0" is $max(p_{channel\ value}) = p_{main\ channel\ value} = 255$ (marked in gray). All values in this real channel in Fig. 5 indicate the severity of the feature in Fig. 4 $a$ at different points on the image (Fig. 1). The dimension of the real channel after the first convolution with a kernel of 5:5 pixels decreases accordingly and is equal to 24:24.

real channel 0 for layer 1

```
0 0 0 0 0 0 0 0 0 0 0 0 0 0 0 0 0 0 0 0 0 0 0 0
0 0 0 0 0 0 0 0 0 0 0 0 42 0 0 0 0 0 0 0 0 0 0 0
0 0 0 0 0 0 0 0 0 0 0 184 42 0 0 0 0 0 0 0 0 0 0 0
0 0 0 0 0 0 0 0 42 184 42 0 0 0 0 0 0 0 0 0 0 0 0 0
0 0 0 0 0 0 0 42 255 42 0 0 0 0 0 0 0 0 0 0 0 0 0 0
0 0 0 0 0 0 0 184 42 0 0 0 0 0 0 0 0 0 0 0 0 0 0 0
0 0 0 0 0 0 42 42 0 0 0 0 0 0 0 0 0 0 0 0 0 0 0 0
0 0 0 0 0 0 42 0 0 0 0 0 0 0 0 0 0 0 0 0 0 0 0 0
0 0 0 0 0 0 42 0 0 0 0 0 0 0 0 0 0 0 0 0 0 0 0 0
0 0 0 0 0 42 0 0 0 0 0 0 0 0 0 0 0 0 0 0 0 0 0 0
0 0 0 0 0 0 0 0 0 0 0 0 0 0 0 0 0 0 0 0 0 0 0 0
0 0 0 0 0 0 0 0 0 0 0 0 0 0 0 0 0 0 0 0 0 0 0 0
0 0 0 0 0 0 0 0 0 0 0 0 0 0 0 0 0 0 0 0 0 0 0 0
0 0 0 0 0 0 0 0 0 0 0 0 0 0 0 0 42 0 0 0 0 0 0 0
0 0 0 0 0 0 0 0 0 0 0 0 0 0 0 113 42 0 0 0 0 0 0 0
0 0 0 0 0 0 0 0 0 0 0 0 0 0 42 184 0 0 0 0 0 0 0 0
0 0 0 0 0 0 0 0 0 0 0 0 0 113 0 0 0 0 0 0 0 0 0 0
0 0 0 0 0 0 0 0 0 0 0 0 0 0 0 0 0 0 0 0 0 0 0 0
0 0 0 0 0 0 0 0 0 0 0 0 0 0 0 0 0 0 0 0 0 0 0 0
0 0 0 0 0 0 0 0 0 0 0 0 0 0 0 0 0 0 0 0 0 0 0 0
0 0 0 0 0 0 0 0 0 0 0 0 0 0 0 0 0 0 0 0 0 0 0 0
0 0 0 0 0 0 0 0 0 0 0 0 0 0 0 0 0 0 0 0 0 0 0 0
```

Fig. 5. Real channel of the first (zero) convolutional layer for the selected image "0", determined based on the kernel in Fig. 4c.

Similarly, based on the displacement b and the resulting kernel (Fig. 4c) for a given feature (Fig. 4a), the real first (zero) channels for all selected 10 images are calculated. In Fig. 6 shows 10 randomly selected images - one image from each class. Each image shows the location of the feature (Fig. 4a), the values of which in the real channels of the first convolutional layer are greater than the value 127.

Fig. 6. Location of the most pronounced feature values in Fig. 4a on all 10 images.

After calculating each value $p^{(layer\_1)}_{channel\ value}$ the value $z_{ij} = 1$ corresponding to this value with coordinates $i$ and $j$ in the common feature channel is reset to zero.

Similarly, we calculate the kernels and channels for the following features in the common feature channel. As a result, we obtain $n_{kan}$ real channels after the first convolution for each selected image (Fig. 7 a). Each channel corresponds to one feature and, accordingly, one kernel.

Fig. 7. Structure of the calculated channels for the first convolutional layer, (*a*) for implementation with image channels, (*b*) for implementation without image channels.

Simultaneously with the calculation of real channels, the general features of the channels (a table of ones and zeros) are also calculated for the second convolutional layer, in which the coordinates of cells with values of 1 will indicate the locations of the features in the real channels of the first convolutional layer. The values of the common feature channel for the second convolutional layer are determined by the following rule:

If $p_{value\ channel,\ i,\ j}$ the channel of the first convolutional layer is greater than 0, then the value of the common feature channel $z_{ij} = 1$, otherwise $z_{ij} = 0$ (Fig. 8a).

In addition, to avoid the appearance of too many channels in the second convolutional layer with similar features, after creating a common channel (Fig. 8a), an algorithm is executed that reduces the number of units in the feature channel. This algorithm removes units in the feature channel that correspond to real channel values with values less than 127, and also removes units in the feature channel that are within 5 pixels of each other. Thus, the number of units in the feature channel is discharged, and only the most significant feature points of the real channels of the first layer remain (Fig. 8).

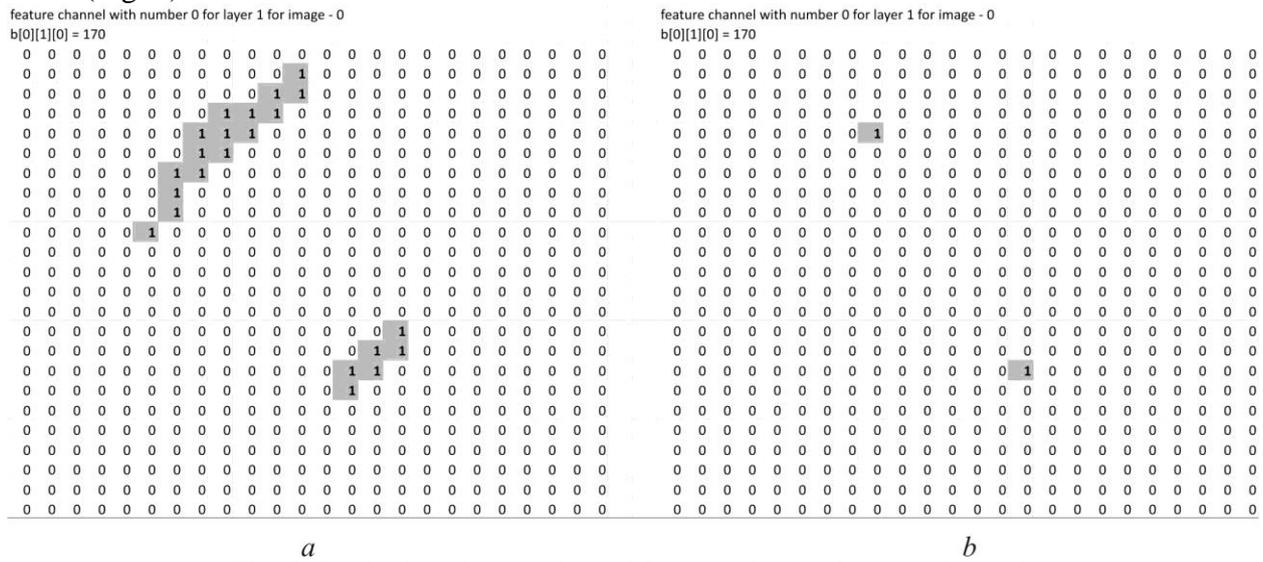

Fig. 8. Reducing the number of features in the feature channel.

In convolutional neural network architectures, at this stage, channel pooling can be performed (which is optional), in which the dimension of the real channel matrices is reduced. It should be noted that if the channels of the first convolutional layer are pooled, then the common feature channels for the second convolutional layer are also pooled (Fig. 9ab). If, for example, channel pooling is performed with a step of 2:2 and with the rule for determining the maximum value in each matrix 2:2 (Fig. 9a), then the dimension of real and common channels is reduced by half (Fig. 9cd).

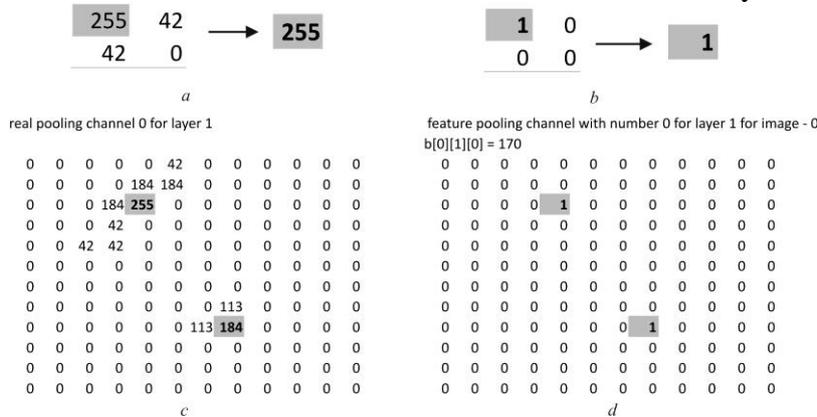

Fig. 9. (*a*) - calculation of the channel pooling based on the maximum value for a real channel, (*b*) - channel pooling calculation for the feature channel, (*c*) – calculated pooling channel for the real channel, (*d*) – calculated pulling channel for the feature channel.

Since each digit in the feature channel of the second convolutional layer must correspond to a feature in the real channel with a table dimension of 5 by 5 cells, then for this purpose the feature channel in Fig. 9 d is first convolved. Convolution is performed according to the rule : the value of the table cell of the common channel after convolution is equal to 1 if in any array *M1[5][5]* of the feature channel there is a central array *M1[2][2]* , in the values of which there is at least at least one unit (Fig. 10*a*). An example of the resulting common channel after convolution is shown in Fig. 10*b*. The dimension of the feature channel is reduced to a dimension of 8 by 8 cells.

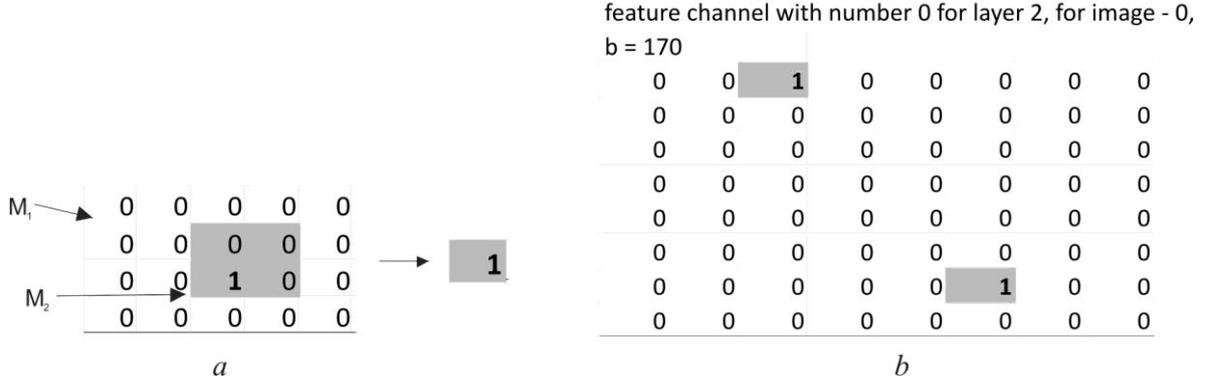

Fig. 10. Convolution of the feature channel for the second convolutional layer, (*a*) selection of a part of the feature channel for convolution, (*b*) feature channel after convolution.

Thus, we obtained the real channels of the first convolutional layer and the feature channels for the second convolutional layer. Based on this data, kernels, offsets and real channels for the second convolutional layer will be calculated.

In this algorithm, channels are calculated for each selected image separately (Fig. 7a). The total number of channels in this case for the first convolutional layer will be equal to:

$$N_{chan} = n_{chan} \cdot N_{img},  \qquad (7)$$

It must be said that the meaning $N_{kan}$ in expression (7) can be reduced to the value $n_{chan}$ if, when calculating the channel, all values of $p_{value\ channel,\ i,j}$ real channels (Fig. 5) are not recorded in separate channels for each image, as shown in Fig. 7*a*, but recorded directly into one real channel (Fig. 7*b*). In this case, after the activation function condition (6) is met, the value $p_{value\ channel,\ i,j}$ will be written to the real channel cell only if the calculated value $p_{value\ channel,\ i,j}$ is greater than the previous value of this cell. Such an implementation will be referred to below as an implementation without calculation of image channels (Fig. 7). At the same time, common image feature channels will be combined in the same way - into one feature channel. In Fig.11 shows examples of the calculated one first (zero) real channel (Fig. 11*a*) and the feature channel (Fig. 11b) for the first convolutional layer with such an implementation.

It must be said that the experiments carried out below showed that computed convolutional networks work both in implementations with image channels (Fig. 7*a*) and in implementations without image channels (Fig. 7*b*). In the case of implementing a computational convolutional neural network without image channels, when creating a feature channel, the algorithm described above, which reduces the number of units in the feature channel, was not used, since the use of this algorithm in an implementation without image channels reduces the number of feature of images too much.

Fig. 11. (*a*) Real channel for the first convolutional layer when implementing a convolutional neural network without image channels, (*b*) Feature channel for the first convolutional layer when implementing a convolutional neural network without image channels.

## 2. COMPUTING THE SECOND CONVOLUTIONAL LAYER

The kernels of the second convolutional layer are calculated based on the previously obtained feature channels of the 2nd layer (Fig. 10*b*) and the real channels of the first layer (Fig. 9*c*). Unlike the channel kernel of the first convolutional layer (Fig. 4*c*), the kernel of the second convolutional layer for one channel is multilayer (Fig. 12*c*). The number of layers of kernel of the second layer is equal to the number of real channels of the first layer. The algorithm for calculating the kernel of the second convolutional layer for one channel sequentially revises the feature channels of the first convolutional layer from channel to channel (Fig. 12*b*). If the *i*-th feature channel of the first convolutional layer contains a value of 1, then a kernel layer is created for the second convolutional layer with a matrix of dimensions 5 by 5 cells, each value of which will be equal to 1. An example is shown in Fig. 12*c*. If the value 0 is found in the i-th feature channel of the first convolutional layer, then, on the contrary, a matrix kernel layer with a dimension of 5 by 5 cells is created, each value of which will be equal to 0 (Fig. 12*c*).

For the second convolutional layer, the bias values *b* were chosen to be 0. Selecting the value *b*=0 due to the multilayer nature of the kernel of the second convolutional layer. Calculated value of *b* according to expression (2) leads to too large a value of *b* and strong decrease in value $p^{(2\ layer)}_{value\ channel}$ and, as a result, significant and informative layers of the kernel and feature in expression (8) are lost.

Based on the obtained kernel values for the first channel of the second convolutional layer and the bias *b*, the values of the real first channel of the second convolutional layer are calculated according to expression (8), where *k* and *l* are the changing horizontal and vertical coordinates on the matrix of the real channel of the first convolutional layer:

$$p^{(2\ layer)}_{value\ channel} = \bar{M}^{(2\ layer)}_{kernel} \cdot \bar{M}^{(1\ layer)}_{img.} - b^{(2\ layer)}_m =$$

$$= \sum_{m=1}^{N^{(1\ layer)}_{chan.}} \sum_{i=1}^{5} \sum_{j=1}^{5} \left( p^{(1\ layer)}_{m,k+i,l+j} \cdot w^{(2\ layer)}_{m,i,j} \right) - b^{(2\ слоя)}_m$$

(8)

In expression (8), the third dimension *m* is number the real channel of the first convolutional layer.

By the same analogy as with calculating the kernel for the first convolutional layer, the values of the resulting kernel (Fig. 12*c*) for the channel of the second convolutional layer are adjusted so that the values of the real channels in the second convolutional layer also do not exceed 255. To do this, it is also calculated $p_{main\ value\ channel}$ according to expression (8) by multiplying the kernel in Fig. 12*c* by the matrix of the feature on the basis of which this kernel was created (for the kernel in Fig. 12*c* this is the layered feature in Fig. 12*a*). Next, we calculate the coefficient *t* using formula (4) and adjust all values $w_{m,i,j}$ of the kernel of the first channel of the second convolutional layer according to expression (9).

$$w_{m,i,j} = w_{m,i,j}/t \qquad (9)$$

Just like for the first convolutional layer, if in the second convolutional layer $p_{main\ value\ channel} < 0$, then this channel is not calculated. This feature is skipped and the transition to the next feature column $z_{kij}$ in Fig. 12*b* in the feature channel with other coordinates *i*, *j* is performed.

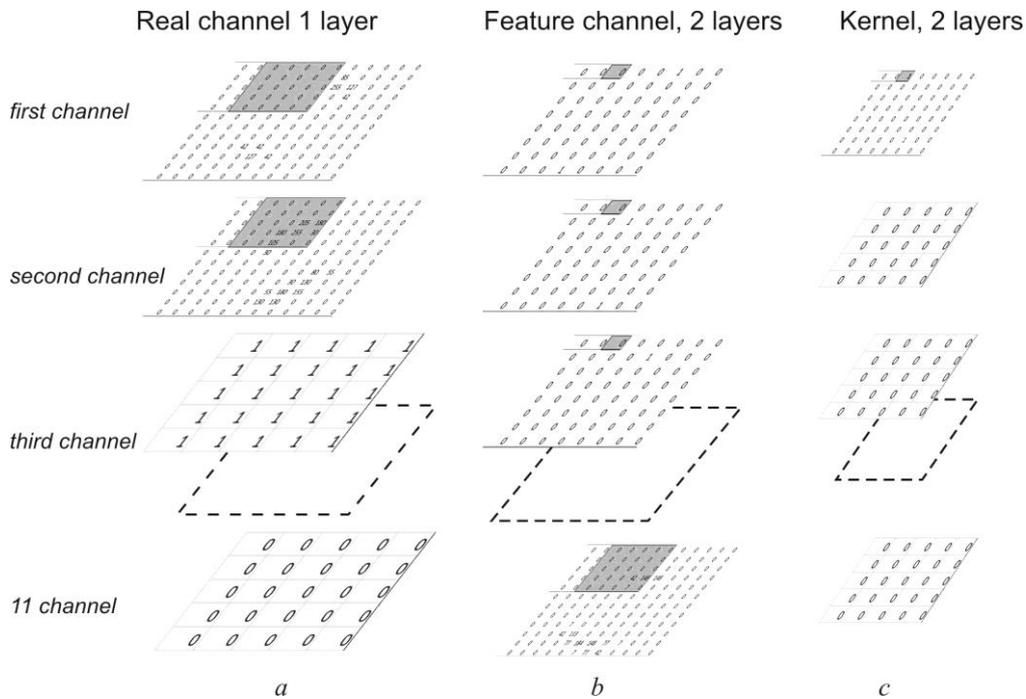

Fig. 12. Calculation of the channel kernel for the second convolutional layer, (*a*) – feature extraction in all real channels in the first convolutional layer, (*b*) – feature extraction in all feature channels in the second convolutional layer, (*c*) – channel kernel of the second convolutional layer without corrections.

After the kernel of the first (zero) channel of the second convolutional layer is calculated, based on this kernel (Fig. 12*c*), the real first channel of the second convolutional layer is calculated separately for all 10 images for the case of implementing a convolutional neural network with image channels (Fig. 7*a*). The activation function of the neuron of the second convolutional layer is determined in the same way as for the first convolutional layer according to expression (6). After calculating each value $p^{(2\ layer)}_{value\ channel,\ i,j}$ in the feature channels the values $z_{kij}$ corresponding to these coordinates *i* and *j* are reset to zero. Examples of calculated real channels for different images of the first (zero) channel are shown in Fig. 13*ab*. When implementing a convolutional network without using image channels, one real first (zero) channel is created (Fig. 7*b*). In this case, the calculated values for different images are written to one channel, provided that the value $p^{(2\ layer)}_{value\ channel}$ written to the cell is greater than the previous value of the channel cell.

| real channel 0 for layer 2 for image 1 | | | | | | | |
|---|---|---|---|---|---|---|---|
| 10 | 107 | 107 | 107 | 107 | 96 | 0 | 0 |
| 21 | 69 | 69 | 69 | 69 | 48 | 0 | 0 |
| 21 | 21 | 21 | 21 | 21 | 0 | 0 | 0 |
| 21 | 21 | 21 | 21 | 21 | 0 | 0 | 0 |
| 21 | 21 | 21 | 21 | 21 | 0 | 0 | 0 |
| 21 | 21 | 21 | 21 | 21 | 0 | 0 | 0 |
| 10 | 10 | 10 | 10 | 10 | 0 | 0 | 0 |
| 0 | 0 | 0 | 0 | 0 | 0 | 0 | 0 |

*a*

| real channel 0 for layer 2 for image 3 | | | | | | | |
|---|---|---|---|---|---|---|---|
| 99 | 176 | 243 | 214 | 165 | 144 | 66 | 0 |
| 165 | 232 | 299 | 269 | 210 | 133 | 66 | 0 |
| 165 | 232 | 299 | 288 | 210 | 133 | 66 | 0 |
| 243 | 310 | 365 | 288 | 210 | 162 | 125 | 58 |
| 302 | 369 | 310 | 232 | 155 | 117 | 80 | 80 |
| 291 | 291 | 232 | 155 | 77 | 51 | 80 | 80 |
| 214 | 214 | 155 | 77 | 10 | 51 | 80 | 80 |
| 136 | 136 | 77 | 10 | 10 | 51 | 80 | 80 |

*b*

Fig. 13. Examples of the calculated first (zero) real channels for the second convolutional layer for (*a*) selected image "1" and (*b*) selected image "3".

In parallel with the calculation of the real first channel, feature channels are also calculated for all 10 images of the third convolutional layer, by analogy with how it was described above for the feature channels of the first convolutional layer. This will be needed if the third convolutional layer is created.

After calculating the first (zero) channel of the second convolutional layer, a transition is made to the next column of features $z_{kij}=1$ in the feature channels with other coordinates *i* and *j* in which there is a value of 1 (Fig. 12*b*). The procedure is repeated with the calculation of kernels for the next channels of the second convolutional layer, the calculation of the next in order real channels for the second convolutional layer and the feature channels for the third convolutional layer until all the features in the feature channels (Fig. 12*b*) of the second convolutional layer are detected. In this implementation of the algorithm, 2 convolutional layers are used for images of the MNIST database. But if necessary, all subsequent convolutional layers are calculated in the same way as for the second convolutional layer, since all subsequent convolutional layers will also have multilayer kernels, as in the second convolutional layer. Fig. 14 shows the locations of all features on all 10 images, corresponding to the first (zero) real channel of the second convolutional layer, the values of which are greater than 127. The features in Fig. 14 correspond to the calculated convolutional neural network implemented with image channels. In Fig. 14 also shows the features of the first (zero) real channel of the second convolutional layer for the input recognized image "2" with sequence number 1 taken from the MNIST test database. From Fig. 14 shows that specifically for the first real channel, the location of the features of the input tested element "2" is most similar to the location of the features in the selected image "2".

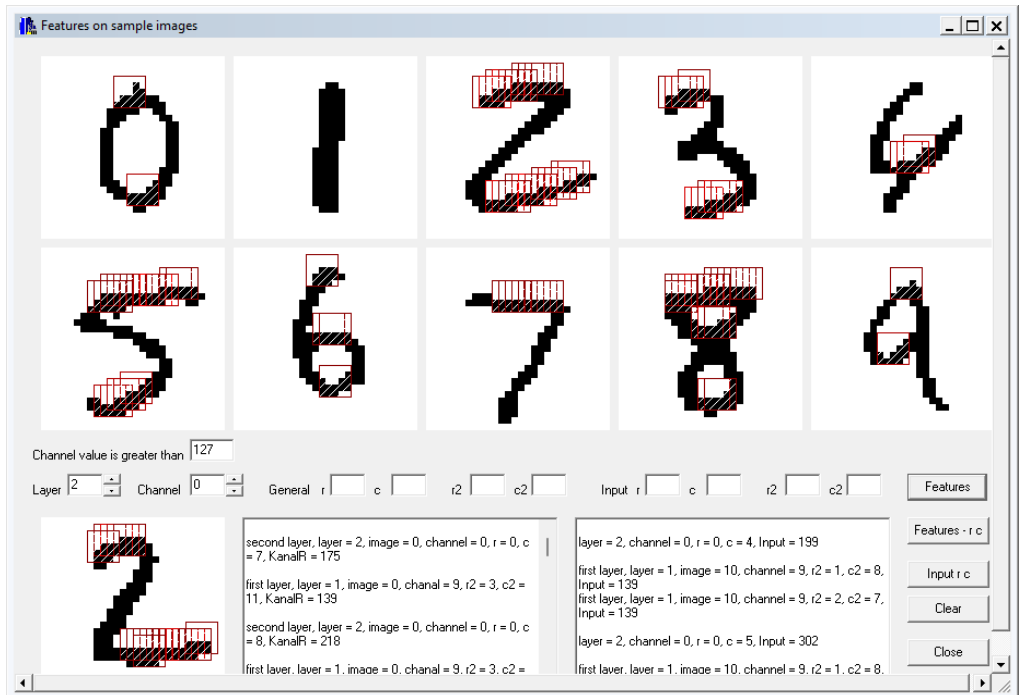

Fig. 14. All features on all 10 images and input image "2" for the first (zero) real channel of the second convolutional layer.

## 3. CALCULATION OF PARAMETERS OF A FULLY CONNECTED NEURAL NETWORK

Based on the calculated last convolutional layer, a three-layer fully connected neural network is created - a perceptron based on metric recognition methods (Fig. 15), the architecture of which was described in [3, 4]. Here, for each output real channel of each selected image of the second convolutional layer, the weight values of the zero layer are calculated. The weight value of each cell of the weight table of the zero layer $w^{(0)}_{k,i,j}$ is calculated by analogy with the physical formula for the electrostatic field strength (10). Weight value $w^{(0)}_{k,i,j}$ with the coordinates of table cell $i, j$ in the $k$-th channel is equal to the highest value of the voltage created by a charge from any one channel cell with cell coordinates $i1, j1$:

$$w^{(0)}_{e,k,i,j} = \frac{p^{(2\,layer)}_{e,k,i1,j1}/1000}{1+(i-i1)^2+(j-j1)^2}, \quad (10)$$

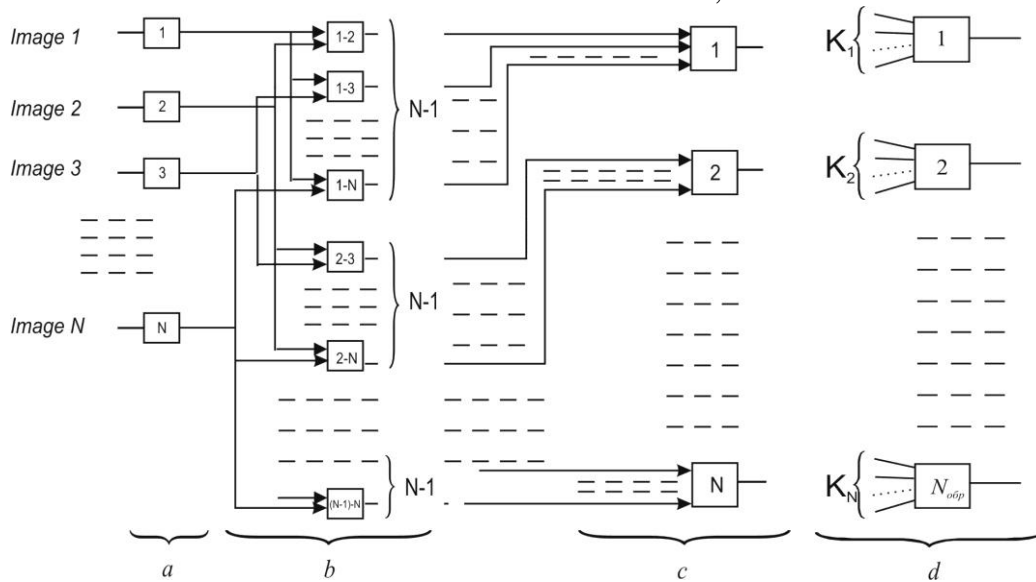

Fig. 15. Three-layer neural network implementing the nearest neighbors method.

| real channel 0 for layer 2 for image 0 | | | | | | | | weights for the zero layer of the neural network for the image - 0 | | | | | | | |
|---|---|---|---|---|---|---|---|---|---|---|---|---|---|---|---|
| | | | | i1 = 4 | | i = 6 | | | | | | | | | |
| 195 | 254 | 254 | 243 | 173 | 58 | 0 | 0 | 0,195 | 0,254 | 0,254 | 0,243 | 0,173 | 0,0865 | 0,0346 | 0,0173 |
| 195 | 243 | 243 | 232 | 162 | 48 | 0 | 0  j1 = 1 | 0,195 | 0,243 | 0,243 | 0,232 | 0,162 | 0,081 | 0,0324 | 0,0162 |
| 147 | 147 | 147 | 136 | 66 | 0 | 0 | 0  j = 2 | 0,147 | 0,147 | 0,147 | 0,136 | 0,081 | 0,054 | 0,027 | 0,0178 |
| 32 | 32 | 32 | 51 | 29 | 29 | 29 | | 0,0735 | 0,0735 | 0,0735 | 0,068 | 0,0535 | 0,0535 | 0,0535 | 0,0385 |
| 21 | 21 | 51 | 117 | 107 | 107 | 107 | 77 | 0,0294 | 0,0294 | 0,0585 | 0,117 | 0,107 | 0,107 | 0,107 | 0,077 |
| 0 | 0 | 29 | 107 | 107 | 107 | 107 | 77 | 0,0147 | 0,0214 | 0,0535 | 0,107 | 0,107 | 0,107 | 0,107 | 0,077 |
| 0 | 0 | 29 | 107 | 107 | 107 | 107 | 77 | 0,0107 | 0,0214 | 0,0535 | 0,107 | 0,107 | 0,107 | 0,107 | 0,077 |
| 0 | 0 | 29 | 107 | 107 | 107 | 107 | 77 | 0,0107 | 0,0214 | 0,0535 | 0,107 | 0,107 | 0,107 | 0,107 | 0,077 |

*a*  *b*

Fig. 16. Creating a table of weights of the zero layer of a fully connected neural network based on the real channel of the second convolutional layer: (*a*) The first (zero) real channel for the image "0" of the second convolutional layer, (*b*) Part of the table of weights of the zero layer for the fully connected neural network.

For example, in Fig. 16 for the real channel of the first image "0" for a cell with coordinates (6, 2), the maximum weight value is calculated using expression (10) as follows:

$$w_{0,0,6,2}^{(0)} = \frac{162/1000}{1+(6-4)^2+(2-1)^2} = \frac{0,162}{6} = 0,027$$

where the largest weight value is created from the channel cell with coordinates (4, 1) with the value $p_{0,0,4,1}^{(2\,layer)} = 162$. It must be said that other expressions could be used to calculate the weight value, for example, it was possible to use the total value of a physical measure - the intensity or potential of an electrostatic field, created from the charges in all cells of the channel, like this was done in [20, 21].

All weight tables for the zero layer are created in a similar way for all images channels (Fig. 17b). The diagram in Fig. 17 corresponds to the scheme for creating a convolutional network with image channels.

In the case of implementing a convolutional neural network without image channels, then, due to the lack of image channels, it will not be possible to create tables of weights of the zero layer, as was described above according to the diagram in Fig. 17. In this case, to create image channels, selected images are sequentially fed to the input of the resulting convolutional neural network (Fig. 18). After submitting each selected image to the input, based on the already obtained kernels, we calculate the channels of the second convolutional layer. Thus, as a result, for these 10 images we obtain output real channels, on the basis of which tables of weights of the zero layer are calculated for 10 images (Fig. 18). It should be noted that in this algorithm, the weight tables of the zero layer were calculated only during the creation of the convolutional neural network and were used only to calculate the weight tables of the first layer of the output fully connected neural network. In testing the convolutional neural network and recognizing the input data, the zero layer weight tables were no longer used.

Based on the obtained tables of weights of the zero layer for each pair of images $n1$ and $n2$ (Fig. 17, 18), the weights are calculated $w_{n1,n2,k,i,j}^{(1)}$ values for the first layer of a fully connected neural network according to the expression:

$$w_{n1,n2,k,i,j}^{(1)} = w_{n1,k,i,j}^{(0)} - w_{n2,k,i,j}^{(0)},$$ (11)

where $k$ is the channel number, $i$ and $j$ are the cell coordinates. Thus, weight tables are calculated for the first layer of the output neural network (Fig. 15b), the number of which is equal to the number of neurons of the first layer of the output fully connected neural network [3]:

$$n = N_{img} \cdot (N_{img} - 1) = 10 \cdot 9 = 90,$$ (12)

where $N_{img}$ is the number of selected images.

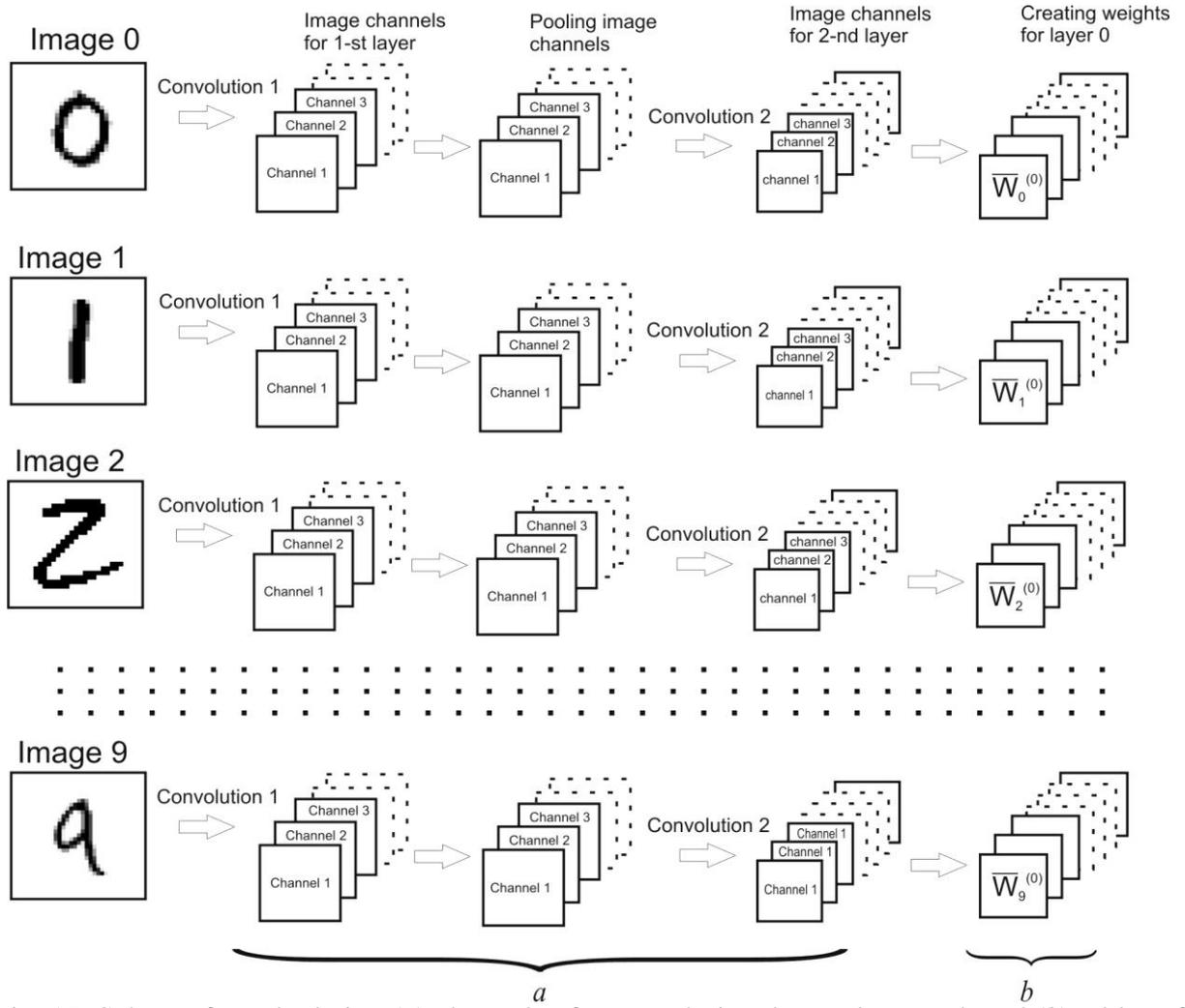

Fig. 17. Scheme for calculating (*a*) channels of a convolutional neural network and (*b*) tables of weights of the zero layer of the output fully connected neural network when implementing a convolutional neural network with image channels.

Next, the threshold values of the neuron of the first layer of the fully connected neural network are calculated. To do this, state values are calculated for all *n* neurons of the first layer of a fully connected neural network using the expression:

$$Sw^{(1)}_{e1,n1,n2} = \sum_{k=1}^{N_{chan}} \sum_{i=1}^{H} \sum_{j=1}^{S} \left( p^{(2\,layer)}_{e1,k,i,j} \cdot w^{(1)}_{n1,n2,k,i,j} \right)$$, (13)

In expression (13) $N_{chan}$ – number of channels in the last (second) convolutional layer, *H* and *S* is the width and height of the channel in the output (in the second) convolutional layer ($H = 8$, $S = 8$, Fig. 13, Fig. 16*a*), *e1* is the number of the image channel in the second convolutional layer (Fig. 17, 18), *n1* and *n2* indicate the number of the weights table of the first layer of the fully connected neural network (11). The threshold value $Wh^{(1)}_{n1,n2}$ of the neuron of the first layer, which performs comparisons between the output image channels *n1* and *n2*, is defined as the average value of the state functions of the neurons of the first layer (13), where *e1* takes the values *n1* and *n2*:

$$Wh^{(1)}_{n1,n2} = -\left( Sw^{(1)}_{n1,n1,n2} + Sw^{(1)}_{n2,n1,n2} \right)/2$$, (14)

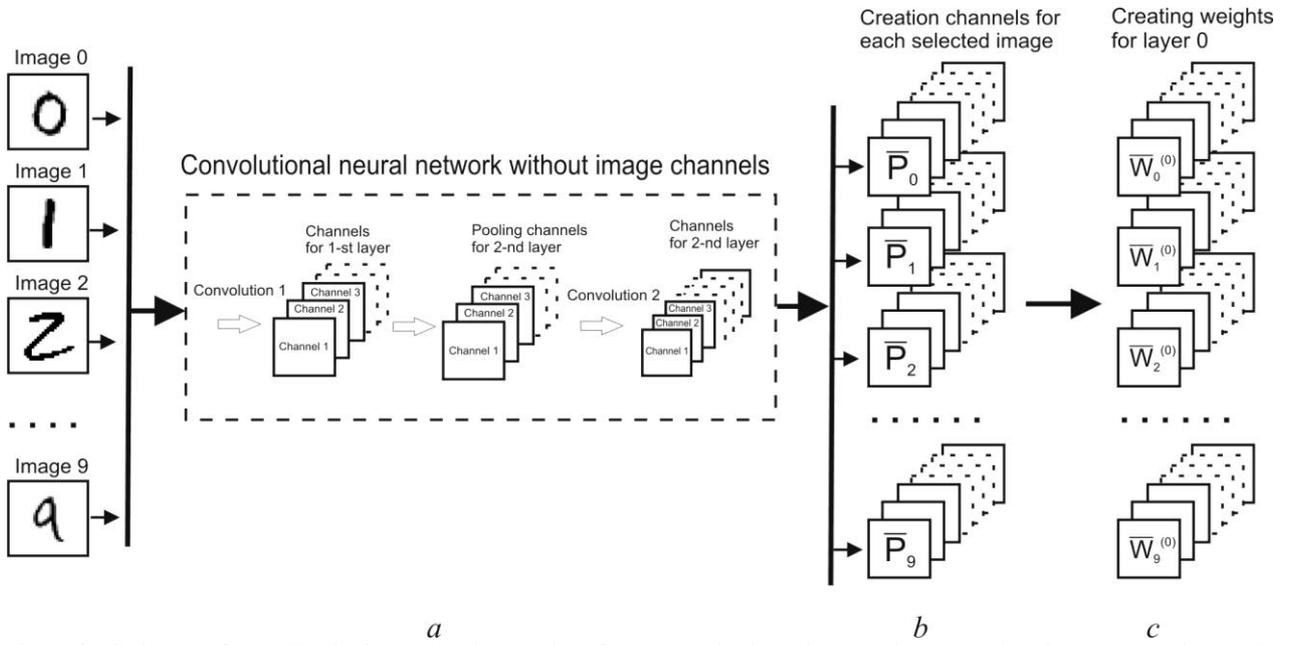

*a*            *b*            *c*

Fig. 18. Scheme for calculating (*a*) channels of a convolutional neural network, (*b*) output channels for each selected image and (*c*) weights of the zero layer of the fully connected neural network when implementing a convolutional neural network without image channels.

Table 1 shows a list of all calculated threshold values of neurons of the first layer of a fully connected neural network, calculated using expression (14).

Table 1. Values of all calculated thresholds of neurons of the first layer of the fully connected neural network.

| | | | | |
|---|---|---|---|---|
| Wh1[0][1] = -9695 | Wh1[2][0] = 5244 | Wh1[4][0] = 21560 | Wh1[6][0] = 24960 | Wh1[8][0] = 18336 |
| Wh1[0][2] = -5244 | Wh1[2][1] = -9377 | Wh1[4][1] = 8273 | Wh1[6][1] = 12613 | Wh1[8][1] = 5566 |
| Wh1[0][3] = -21297 | Wh1[2][3] = -16925 | Wh1[4][2] = 16866 | Wh1[6][2] = 21795 | Wh1[8][2] = 15057 |
| Wh1[0][4] = -21560 | Wh1[2][4] = -16866 | Wh1[4][3] = 833 | Wh1[6][3] = 5556 | Wh1[8][3] = -1592 |
| Wh1[0][5] = -26932 | Wh1[2][5] = -20909 | Wh1[4][5] = -4733 | Wh1[6][4] = 4520 | Wh1[8][4] = -2488 |
| Wh1[0][6] = -24960 | Wh1[2][6] = -21795 | Wh1[4][6] = -4520 | Wh1[6][5] = -210 | Wh1[8][5] = -7821 |
| Wh1[0][7] = -31905 | Wh1[2][7] = -27718 | Wh1[4][7] = -10859 | Wh1[6][7] = -5851 | Wh1[8][6] = -7682 |
| Wh1[0][8] = -18336 | Wh1[2][8] = -15057 | Wh1[4][8] = 2488 | Wh1[6][8] = 7682 | Wh1[8][7] = -14054 |
| Wh1[0][9] = -20449 | Wh1[2][9] = -15953 | Wh1[4][9] = 1210 | Wh1[6][9] = 5221 | Wh1[8][9] = -1545 |
| Wh1[1][0] = 9695 | Wh1[3][0] = 21297 | Wh1[5][0] = 26932 | Wh1[7][0] = 31905 | Wh1[9][0] = 20449 |
| Wh1[1][2] = 9377 | Wh1[3][1] = 7676 | Wh1[5][1] = 13799 | Wh1[7][1] = 19396 | Wh1[9][1] = 8243 |
| Wh1[1][3] = -7676 | Wh1[3][2] = 16925 | Wh1[5][2] = 20909 | Wh1[7][2] = 27718 | Wh1[9][2] = 15953 |
| Wh1[1][4] = -8273 | Wh1[3][4] = -833 | Wh1[5][3] = 5377 | Wh1[7][3] = 11052 | Wh1[9][3] = -140 |
| Wh1[1][5] = -13799 | Wh1[3][5] = -5377 | Wh1[5][4] = 4733 | Wh1[7][4] = 10859 | Wh1[9][4] = -1210 |
| Wh1[1][6] = -12613 | Wh1[3][6] = -5556 | Wh1[5][6] = 210 | Wh1[7][5] = 5618 | Wh1[9][5] = -5892 |
| Wh1[1][7] = -19396 | Wh1[3][7] = -11052 | Wh1[5][7] = -5618 | Wh1[7][6] = 5851 | Wh1[9][6] = -5221 |
| Wh1[1][8] = -5566 | Wh1[3][8] = 1592 | Wh1[5][8] = 7821 | Wh1[7][8] = 14054 | Wh1[9][7] = -12178 |
| Wh1[1][9] = -8243 | Wh1[3][9] = 140 | Wh1[5][9] = 5892 | Wh1[7][9] = 12178 | Wh1[9][8] = 1545 |

Fig. 19*a* shows a fragment of the weights of the second layer for the output fully connected neural network, where each *k*-th line of digits of 1 and 0 corresponds to one *k*-th neuron of the second layer of the output perceptron. Each *i*-th digit in the line corresponds to the weight value for the *i*-th connection of the *k*-th neuron of the second layer, connecting the given *k*-th neuron of the second layer with the *i*-th neuron of the first layer (Fig. 15*b*). Accordingly, the number of digits in each line is equal to the number of neurons in the first layer, $n^{(1)} = 90$.

Threshold value of the second layer neuron (bias) determined by the formula:

$$Wh2 = -(N_{img} - 1) = -(10 - 1) = -9 \qquad (15)$$

where $N_{img}$ is the number of selected images ($N_{img}$ = 10). The number of neurons in the second layer is equal to the number of selected images $n^{(2)} = 10$.

Fig. 19. (*a*) Values of connection weights and thresholds of neurons of the second layer for the output fully connected neural network, (*b*) Values of connection weights and thresholds of neurons of the third layer for the output fully connected neural network.

All weights of the third layer of a fully connected neural network are shown in Fig. 19*b*. The neuron of the third layer of a fully connected neural network combines the outputs of the second layer of the neural network, corresponding to the selected images of one class. Since in this example one image from each recognized class is used, the use of the third layer is accordingly optional. The number of neurons in the third layer is equal to the number of recognized classes $n^{(3)} = 10$. The values of all thresholds of the third layer are $Wh3 = 0$ (Fig. 19*b*).

## 4. EXPERIMENTS WITH A COMPUTED CONVOLUTIONAL NEURAL NETWORK BASED ON MNIST

Based on the algorithms described above for creating a computed convolutional neural network, a software application was created in the C ++ Builder environment (Fig. 20). Based on this application, a number of experiments were carried out with the creation of a computed convolutional neural network and testing the performance of the computed neural network based on the MNIST test database with a set of 1000 tested images of numbers (Fig. 20.1). In all experiments, two convolutional layers were created for the neural network (Fig. 20.3) with a 5 by 5 pixel kernel (Fig. 20.4). A combo list (Fig. 20.2) of 10 images (one image from each class) randomly selected from the MNIST database was used as selected images (Fig. 6). When calculating channels for a convolutional neural network, only those channels whose feature channels had more than four units were saved (Fig. 20.7). This was done in order to exclude atypical (noise) features of the selected images for a given class. In addition to the convolutional layers, a fully connected three-layer neural network was also calculated, as described above. As a result, in the experiments below, 5 or 6 (taking into account the pooling channel) layer neural networks were created.

In the first experiment, a convolutional neural network with image channels was created according to the scheme in Fig. 17. In this experiment, a pooling layer with a compression size of 2 by 2 pixels was also created between the two convolutional layers (Fig. 9, 20.5). The value of the threshold coefficient *K* in (2) was taken equal to 40% (Fig. 20.6).

As a result, for the first experiment, a convolutional neural network was analytically created with 11 channels in the first convolutional layer and with 46 channels in the second convolutional layer (Fig. 20.8) in each image channel. The total number of channels for the first convolutional layer was 11*10=110 channels, and for the second convolutional layer was 46*10=460 channels. The number of calculated kernels was 11 kernels for the first convolutional layer and 46 kernels for the second convolutional layer. The creation time of this neural network (Fig. 20.9) was less than 1 second ($t_{creation} = 0.391$ s.). The results of testing the resulting neural network on the MNIST test database showed that the resulting neural network successfully recognized 527 images of numbers out of 1000 possible images of the MNIST test base, which ultimately amounted to 52% recognition performance (Fig. 20.10).

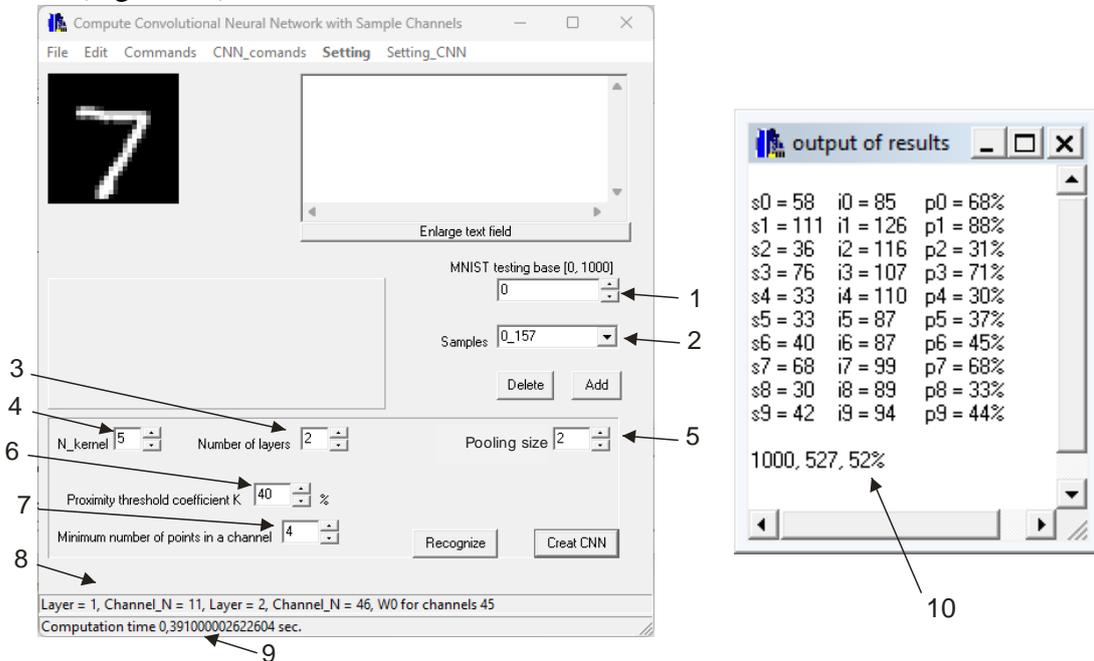

Fig. 20. Creation of a neural network and testing the performance of the resulting neural network based on the MNIST test database implemented with image channels and with a pooling layer.

As a result, for the first experiment, a convolutional neural network was analytically created with 11 channels in the first convolutional layer and with 46 channels in the second convolutional layer (Fig. 20.8) in each image channel. The total number of channels for the first convolutional layer was 11*10=110 channels, and for the second convolutional layer was 46*10=460 channels. The number of calculated kernels was 11 kernels for the first convolutional layer and 46 kernels for the second convolutional layer. The creation time of this neural network (Fig. 20.9) was less than 1 second ($t_{creation} = 0.391$ s.). The results of testing the resulting neural network on the MNIST test database showed that the resulting neural network successfully recognized 527 images of numbers out of 1000 possible images of the MNIST test base, which ultimately amounted to 52% recognition performance (Fig. 20.10).

In the second experiment, a convolutional neural network was also created with image channels and two convolutional layers, but without using a pooling layer (Fig. 21.1). The value of the threshold coefficient in (2) was also taken to be equal to $K = 40\%$. Thus, a convolutional neural network was obtained with 11 channels of the first convolutional layer and 93 channels of the second convolutional layer (Fig. 21.3). As a result, the total number of channels for the first convolutional layer was 11*10=110 channels, and for the second convolutional layer it was 93*10=930 channels. The number of calculated kernels was 11 for the first convolutional layer and 93 kernels for the second convolutional layer. The creation time of this neural network was $t_{creation} = 9.218$ s. (Fig. 21.2). It should be noted that the significant difference in the time of creating a convolutional neural network in the first and second experiments is due to the use of a pooling layer in the first experiment. Using a pooling layer in the first experiment reduces the size of the channel matrix by half, and also leads to a reduction in the channels in the second convolutional layer and, in general, leads to a significant reduction in the computation time of the convolutional neural network. On the other hand, the results of testing the calculated convolutional neural network in the second experi-ment (without a pooling layer) showed that out of 1000 images of digits in the MNIST test database, 583 were accurately recognized, which amounted to 58% of the recognition performance (Fig. 21.4). This is 6% more than in the first experiment.

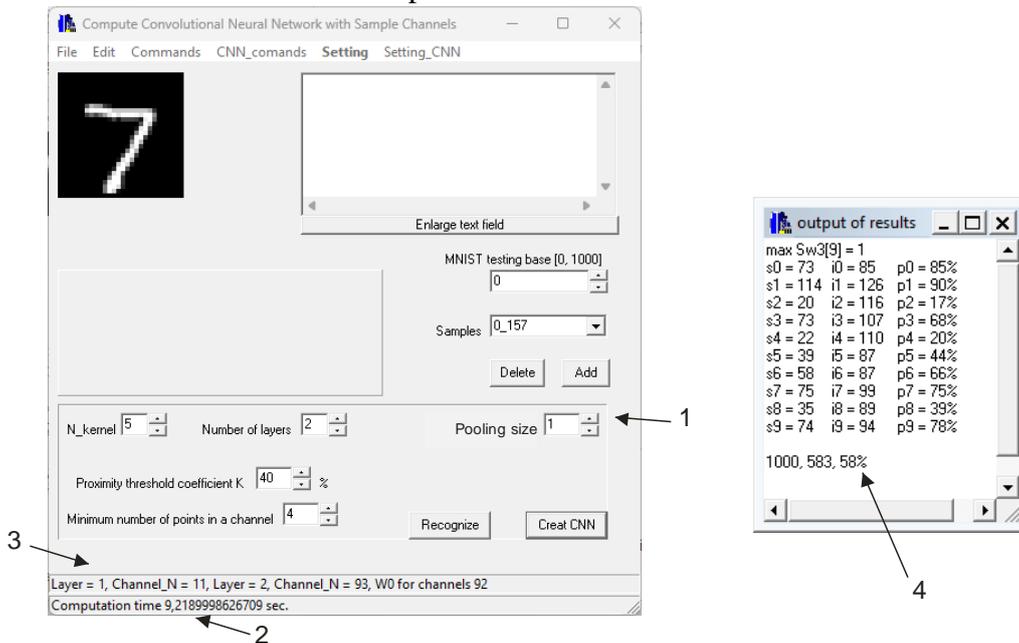

Fig.21. Creating a neural network and checking the performance of the resulting neural network based on the MNIST test base implemented with image channels without using a pooling layer.

Convolutional neural network was created without using image channels (according to the diagram in Fig. 18). The convolutional neural network was created with two convolutional layers without pooling layer. The value of the threshold coefficient *K* in (2) was taken equal to 30% (Fig. 22). As a result, a neural network was calculated with 20 channels in the first convolutional layer and 62 channels in the second convolutional layer (Fig. 22.1). The creation time of this neural network was $t_{creation}$ = 9.204 s. (Fig. 22.2). The results of testing the calculated convolutional neural network on the MNIST test base showed that out of 1000 images of the MNIST test base, 553 were recognized correctly, which amounted to 55% recognition performance (Fig. 22.3).

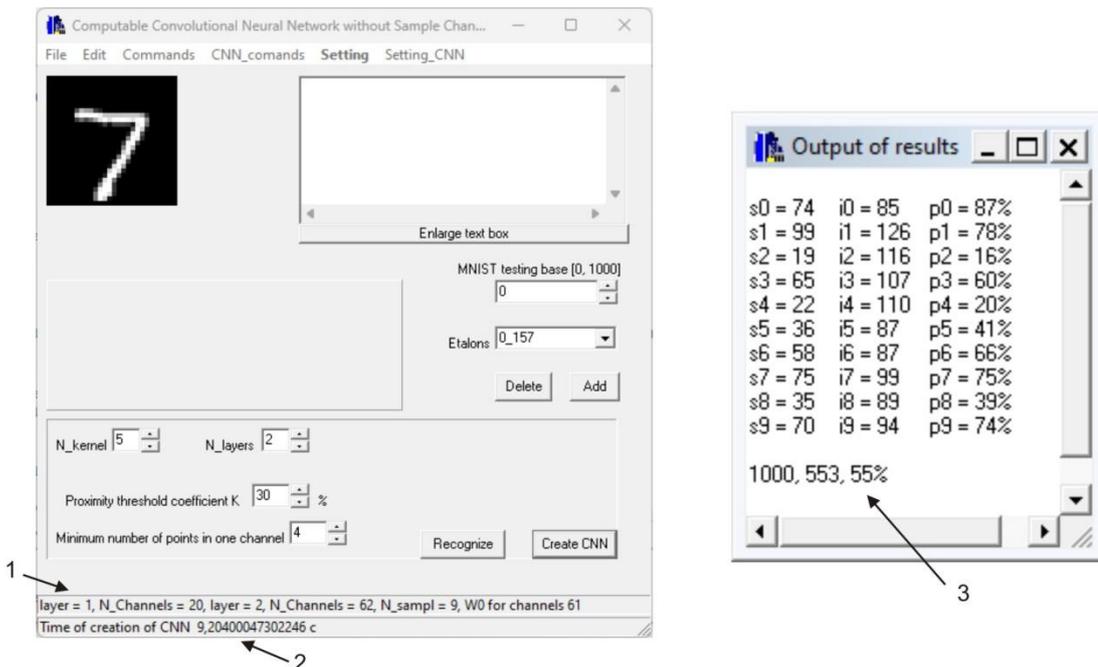

Fig.22. Creating a neural network and testing the performance of the resulting neural network based on the MNIST test base implemented without image channels and without using a pooling layer.

## 5. CONCLUSION

Thus, in this work it was shown that the values of the weights of a convolutional network can be calculated analytically, which allows you to immediately create a workable convolutional neural network. It was also shown that the number of channels in convolutional layers is also determined by the algorithm and depends on the number of features in the selected images. Experiments have shown that the time for creating and analytically calculating a convolutional neural network is very short and amounts to fractions of a second or a minute. Experiments also showed that a small number of selected images is required to compute a convolutional neural network. Even when using only one image for each class, the performance of the neural network is more than 50%.

At the same time, based on the results of experiments in previous works [3, 4], we can also say that the results obtained in this work are not final. Thus, an increase in the number of selected images also increases the effectiveness of the neural network. In addition, after analytically calculating the values of the weights, the neural network can be additionally trained with gradient descent algorithms, for example, the backpropagation algorithm. And as was shown in [3, 4], this training is much faster and requires a smaller training data set.

Note also that the ability to quickly calculate the value of the weights can allow the selection of the most optimal values for analytically indeterminable values and parameters of the neural network, such as $K$ or the dimension of the kernel of a convolutional neural network. The determination of these values can be realized, for example, by enumerating a certain set of discrete values of these parameters with repeated calculations of the weights and checking on test bases. Since the analytical calculation of weights is performed very quickly, such an enumeration will also be performed quickly, compared to traditional schemes for selecting neural network parameters with the need to constantly retrain the neural network after each change in the network parameter. Considering the length of time it takes to train convolutional neural networks, this takes significantly longer.

In general, the proposed technology can speed up the procedure for creating and training convolutional neural networks, and can also allow the neural network to be trained with smaller training data sets. The proposed technology also allows us to better understand the purpose of the internal contents of a trained convolutional neural network, the purpose of individual neurons, weights, etc.